\documentclass[11pt,a4paper]{article}
\usepackage[hyperref]{eacl2021}
\usepackage{times}
\usepackage{latexsym}
\usepackage{xcolor}
\usepackage{pbox}
\usepackage[normalem]{ulem} 
\newcommand\hly{\bgroup\markoverwith
  {\textcolor{yellow!15}{\rule[-.5ex]{2pt}{2.5ex}}}\ULon}
 \newcommand\hlb{\bgroup\markoverwith
  {\textcolor{blue!10}{\rule[-.5ex]{2pt}{2.5ex}}}\ULon}
\usepackage{longtable}
\usepackage[warn]{textcomp}
\usepackage{silence}
\usepackage{scalerel}
\def\shef{\scalerel*{\includegraphics{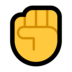}}{\textrm{\textbigcircle}}}
\def\buaa{\scalerel*{\includegraphics{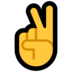}}{\textrm{\textbigcircle}}}
\def\open{\scalerel*{\includegraphics{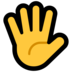}}{\textrm{\textbigcircle}}}
\usepackage{microtype}

\usepackage{CJKutf8}
\newcommand{\Chinese}[1]{\begin{CJK*}{UTF8}{gbsn}{#1}\end{CJK*}}
\newcommand{\TradChinese}[1]{\begin{CJK*}{UTF8}{bsmi}{#1}\end{CJK*}}

\usepackage{enumitem}
\setitemize{noitemsep,topsep=0pt,parsep=0pt,partopsep=0pt}
\usepackage{xspace}
\usepackage{xcolor}
\usepackage{multirow}
\usepackage{tabularx}
\usepackage{booktabs}
\usepackage{arydshln}
\usepackage{subcaption}
\usepackage{graphicx}
\usepackage{float}
\usepackage{cleveref}
\crefname{section}{§}{§§}

\newcommand{\HS}{\textsc{HistSumm}\xspace}
\newcommand{\de}{\textsc{de}\xspace}
\newcommand{\zh}{\textsc{zh}\xspace}
\newcommand{\en}{\textsc{en}\xspace}

\aclfinalcopy 
\def\aclpaperid{1203} 

\newcolumntype{P}[1]{>{\centering\arraybackslash}p{#1}}
\newcolumntype{Y}{>{\centering\arraybackslash}X}

\title{Summarising Historical Text in Modern Languages}

\author{Xutan Peng\textsuperscript{\shef} \hspace{4mm}  Yi Zheng\textsuperscript{\buaa} \hspace{4mm}  Chenghua Lin\textsuperscript{\shef}\thanks{~~Chenghua Lin is the corresponding author.}~ \hspace{4mm}  Advaith Siddharthan\textsuperscript{\open}\\
  \textsuperscript{\shef}Department of Computer Science, The University of Sheffield, UK \\
  \textsuperscript{\buaa}School of Computer Science and Engineering, Beihang University, China \\ 
  \textsuperscript{\open}Knowledge Media Institute, The Open
University, UK \\
  \texttt{\{x.peng, c.lin\}@shef.ac.uk} \hspace{4mm} \texttt{zhengyi53@buaa.edu.cn} \\
  \texttt{advaith.siddharthan@open.ac.uk}
}

\date{}

\begin{document}
\maketitle
\begin{abstract}
We introduce the task of historical text summarisation, where documents in historical forms of a language are summarised in the corresponding modern language. This is a fundamentally important routine to historians and digital humanities researchers but has never been automated. We compile a high-quality gold-standard text summarisation dataset, which consists of historical German and Chinese news from hundreds of years ago summarised in modern German or Chinese. Based on cross-lingual transfer learning techniques, we propose a summarisation model that can be trained even with no cross-lingual (historical to modern) parallel data, and further benchmark it against state-of-the-art algorithms.
We report automatic and human evaluations that distinguish the historic to modern language summarisation task from standard cross-lingual  summarisation (i.e., modern to modern language), highlight the distinctness and value of our dataset, and demonstrate that our transfer learning approach outperforms standard cross-lingual benchmarks on this task.
\end{abstract}

\section{Introduction}
The process of text summarisation is fundamental to research into  history, archaeology, and digital humanities~\citep{hist-research-1}. Researchers can better gather and organise information and share knowledge by first identifying the key points in historical documents. However, this can cost a lot of time and effort. On one hand, due to cultural and linguistic variations over time, interpreting historical text can be a challenging and energy-consuming process, even for those with specialist training~\cite{gray2011language}. To compound this, historical archives can contain narrative documents on a large scale, adding to the workload of manually locating important elements~\citep{hist-research-2}.  To reduce these burdens, specialised software has been developed recently, such as MARKUS~\citep{MARKUS} and DocuSky~\citep{DocuSky}. These toolkits aid users in managing and annotating documents but still lack functionalities to automatically process texts at a semantic level.

Historical text summarisation can be regarded as a special case of cross-lingual summarisation~\citep{leuski2003cross,orasan-chiorean-2008-evaluation, jointly-acl20}, a long-standing research topic whereby summaries are generated in a target language from documents in different source languages. However, historical text summarisation posits some unique challenges. 
Cross-lingual (i.e., across historical and modern forms of a language) corpora are rather limited~\citep{gray2011language} and therefore historical texts cannot be handled by traditional cross-lingual summarisers, which require cross-lingual supervision or at least large summarisation datasets in both languages~\citep{jointly-acl20}. Further, language use evolves over time, including {vocabulary and word spellings} and meanings~\citep{hist-research-2}, and historical collections can span hundreds of years. Writing styles also change over time. For instance, while it is common for today's news stories to present important information in the first few sentences, a pattern exploited by modern news summarisers~\citep{pg}, this was not the norm in older times~\citep{news-intro}.

\begin{table*}[!ht]
\centering 
\bgroup
\def\arraystretch{1.24}
\resizebox{0.97\textwidth}{!}{\begin{tabular}{{p{1.8cm}p{17.5cm}}}
\toprule
\de & \textnumero 34 \\ \midrule
Story & Jhre Königl. Majest. befinden sich noch vnweit Thorn / ... / dahero zur Erledigung Hoffnung gemacht werden will. \newline \textit{\textcolor{darkblue}{(Their Royal Majesties are still not far from Torn, ... , therefore completion of the hope is desired.)}} \\ 
Summary & Der Krieg zwischen Polen und Schweden dauert an. Von einem Friedensvertrag ist noch nicht der Rede. \newline \textit{\textcolor{darkblue}{(The war between Poland and Sweden continues. There is still no talk on the peace treaty.)}} 
\\\midrule
\zh & \textnumero 7 \\ \midrule
Story &  \Chinese{有脚夫小民， 三四千名集众围绕马监丞衙门，...， 冒火突入， 捧出敕印。} \newline
\textit{\textcolor{darkblue}{(Three to four thousand porters gathered around Majiancheng Yamen (a government office), ..., rushed into fire and salvaged the authority's seal.)}} \\ 
Summary & \Chinese{小本生意免税条约未能落实，小商贩被严重剥削，以致百姓聚众闹事并火烧衙门，造成多人伤亡。王炀抢救出公章。} \newline \textit{\textcolor{darkblue}{(The tax-exemption act for small businesses was not well implemented and small traders were terribly exploited, leading to riot and arson attack on Yamen with many casualties. Yang Wang salvaged the authority's seal.)}} \\ \bottomrule
\end{tabular}}
\egroup
\caption{Examples from our \textsc{HistSumm} dataset.} 
\label{tab:example}
\end{table*}

In this paper, we address the long-standing need for historical text summarisation through machine summarisation techniques for the first time. We consider the German{\scriptsize$|$\de} and Chinese{\scriptsize$|$\zh} languages, selected for the following reasons. First, they both have rich textual heritage and accessible (monolingual) training resources for historical and modern language forms. Second, they serve as outstanding representatives of two distinct writing systems (\de for alphabetic and \zh for ideographic languages), and investigating them can lead to generalisable insights for a wide range of other languages. Third, we have access to linguistic experts in both languages, for composing high-quality gold-standard modern-language summarises for \de and \zh news stories published hundreds of years ago, and for evaluating the output of machine summarisers. 

In order to tackle the challenge of a limited amount of resources available for model training (e.g., we have summarisation training data only for the monolingual task with modern languages, and very limited parallel corpora for modern and historical forms of the languages), we propose a transfer-learning-based approach which can be bootstrapped even without cross-lingual supervision. To our knowledge, our work is the first to consider the task of historical text summarisation. As a result, there are no directly relevant methods to compare against. We instead implement two state-of-the-art baselines for standard cross-lingual summarisation, and conduct extensive automatic and human evaluations to show that our proposed method yields better results.  Our approach, therefore, provides a strong baseline for future studies on this task to benchmark against.

The contributions of our work are three-fold: (1) we propose a hitherto unexplored and challenging task of historical text summarisation; (2) we construct a high-quality summarisation corpus for historical \de and \zh, with modern \de and \zh summaries by experts, to kickstart research in this field; and (3) we propose a model for historical text summarisation that does not require parallel supervision and provides a validated high-performing baseline for future studies.
We release our code and data at
\url{https://github.com/Pzoom522/HistSumm}.

\section{Related Work}\label{sec:rw}

\paragraph{Processing historical text.} Early NLP studies for historical documents focus on spelling normalisation~\citep{hist-nlp}, machine translation~\citep{hist-mt}, and sequence labelling applications, e.g., part-of-speech tagging~\citep{pos} and named entity recognition~\citep{ner}. Since the rise of neural networks, a broader spectrum of applications such as sentiment analysis~\citep{sentiment}, information retrieval~\citep{ie}, and relation extraction~\citep{knowledge} have been developed. 

We add to this growing literature in two ways.
First, much of the work on historical text processing is focused on English{\scriptsize$|$\en}, and work in other languages is still relatively unexplored~\citep{hist-nlp, non-en}. Second, the task of historical text summarisation has never been tackled before, to the best of our knowledge. A lack of non-\en annotated historical resources is a key reason {for the former, and for the latter, resources do not exist in any language.} We hope to spur research on historical text summarisation and in particular for non-\en languages through this work.

\paragraph{Cross-lingual summarisation.} The traditional strands of cross-lingual text summarisation systems design pipelines which learn to translate and summarise separately~\citep{leuski2003cross,orasan-chiorean-2008-evaluation}. However, such paradigms suffer from the error propagation problem, i.e., errors produced by upstream modules may accumulate and degrade the output quality~\citep{attend-acl20}. In addition, parallel data to train effective translators is not always accessible~\citep{jointly-acl20}. 
Recently, end-to-end methods have been applied to alleviate this issue. The main challenge for this research direction is the lack of {direct corpora}, leading to attempts such as zero-shot learning~\citep{zeroshot}, multi-task learning~\citep{multitask}, and transfer learning~\citep{jointly-acl20}. Although training requirements have been relaxed by these methods, our extreme setup with summarisation data only available for the target language and very limited parallel data, has never been visited before.

\section{\textsc{HistSumm} Corpus}

\subsection{Dataset Construction}\label{ssec:annotation}

In history and digital humanities research, summarisation is most needed when analysing documentary and narrative text such as news, chronicles, diaries, and memoirs~\citep{hist-research-1}. Therefore, for \de we picked the GerManC dataset~\citep{GerManC}, which contains Optical Character Recognition (OCR) results of \de newspapers from the years 1650--1800. We randomly selected 100 out of the 383 news stories for manual annotation.
For \zh, we chose \Chinese{『万历邸抄』} (\textit{Wanli Gazette}) as the data source, a collection of news stories from the Wanli period of Ming Dynasty (1573--1620). However, there are no machine-readable versions of Wanli Gazette available; worse still, the calligraphy copies (see Appendix~\ref{app:wanli}) are unrecognisable even for non-expert humans, making the OCR technique inapplicable. Therefore, we performed a thorough literature search on over 200 related academic papers and manually retrieved 100 news texts\footnote{Detailed references are included in the `source' entries of \zh \HS's metadata.}.

To generate summaries in the respective modern language for these historical news stories, we recruited two experts with degrees in Germanistik and Ancient Chinese Literature, respectively. They were asked to produce summaries in the style of \de MLSUM~\citep{mlsum} and \zh LCSTS~\citep{lcsts}, whose news stories and summaries are crawled from the Süddeutsche Zeitung website and posts by professional media on the Sina Weibo platform, respectively. 
The annotation process turned out to be very effort-intensive: for both languages, the experts spent at least 20 minutes in reading and composing a summary for one single news story. The accomplished corpus of 100 news stories and expert summaries in each language, namely \textsc{HistSumm} (see examples in Tab.~\ref{tab:example}), were further examined by six other experts for quality control (see details in \cref{ssec:exp-human}).

\subsection{Dataset Statistics}\label{ssec:stats}

\begin{figure}[!t]
  \centering
  \includegraphics[trim={0 0.4cm 0 0.3cm}, clip, width=\columnwidth]{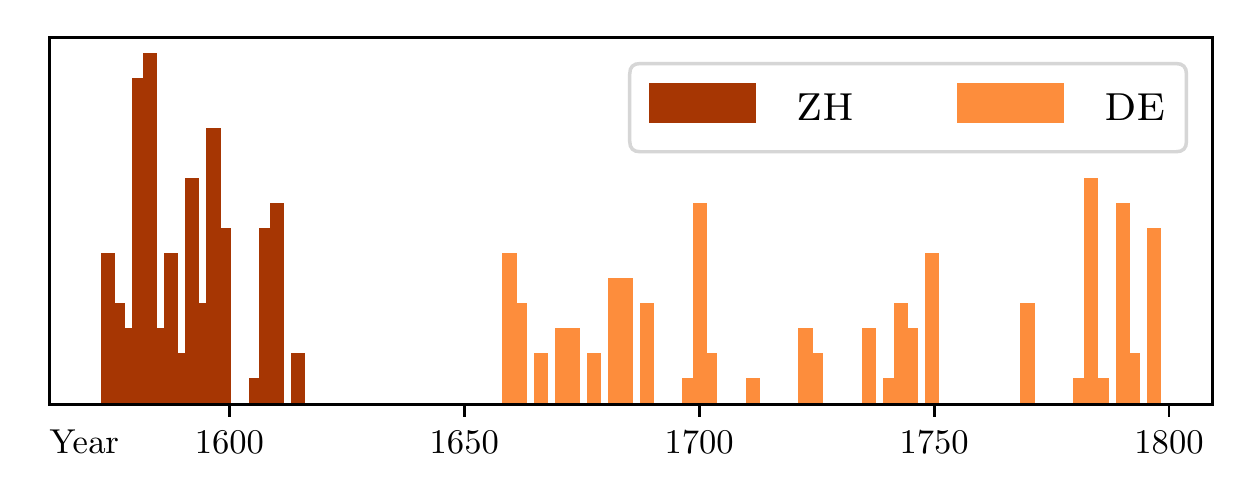}
\caption{Publication time of \textsc{HistSumm} stories.}
\label{fig:year}
\end{figure}

\paragraph{Publication time.}~~As visualised in Fig.~\ref{fig:year}, the publication time of \de and \zh \HS stories exhibits distinguished patterns. Oldness is an important indicator of the domain and linguistic gaps~\citep{hist-research-2}. Considering news in \zh \HS is on average 137 years older than its \de counterpart, such gaps can be expected to be greater. 
On the other hand, \de \HS stories cover a period of 150 years, compared to just 47 years for \zh, indicating the potential for greater linguistic and cultural variation within the \de corpus. 

\begin{figure}[!t]
  \centering
  \includegraphics[trim={0 .5cm 0 .2cm}, clip, width=\columnwidth]{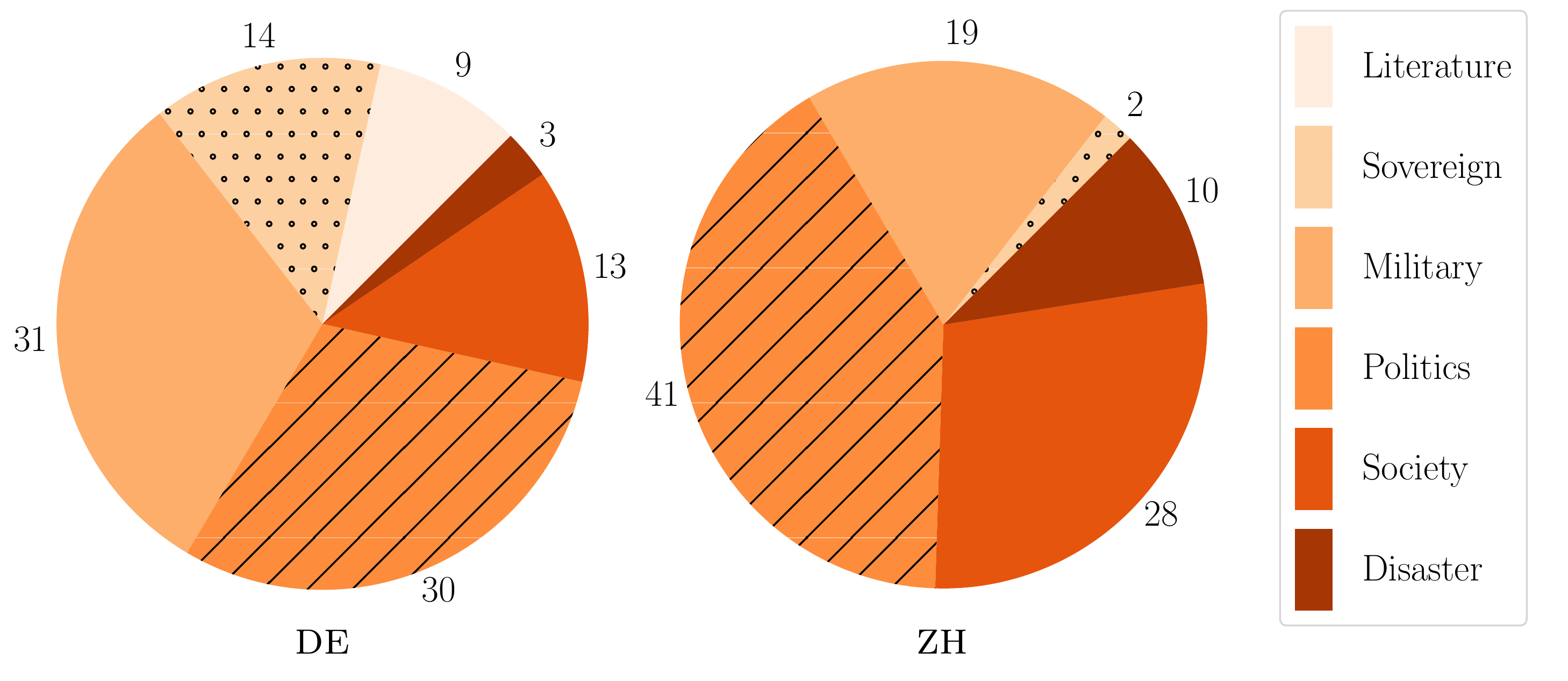}
\caption{Topic composition of \HS.}
\label{fig:topic}
\end{figure}

\begin{table}[!t]\small
\centering 
\begin{tabular}{{p{1.4cm}|P{1.5cm}P{1.5cm}|P{1.5cm}P{1.5cm}}}
\toprule 
& \multicolumn{2}{c|}{\de (word-level)} & \multicolumn{2}{c}{\zh (character-level)} \\ \midrule
 & \textsc{HistSumm} & MLSUM & \textsc{HistSumm} & LCSTS\\ \midrule
$L_{\mathrm{story}}$ & 268.1 & 570.6  & 114.5 & 102.5\\
$L_{\mathrm{summ}}$ & 18.1 & 30.4  & 28.2 & 17.3 \\
$CR$ (\%) & 6.8 & 5.3  & 24.6 & 16.9 \\ \bottomrule
\end{tabular}
\caption{Comparisons of mean story length ($L_{\mathrm{story}}$), summary length ($L_{\mathrm{summ}}$), and compression rate ($CR={L_{\mathrm{summ}}}/{L_{\mathrm{story}}}$) for summarisation datasets.}
\label{tab:dataset_stats}
\end{table}

\paragraph{Topic composition.} 
For a high-level view of \textsc{HistSumm}'s content, we asked experts to manually classify all news stories into six categories (shown in Fig.~\ref{fig:topic}).  We see that the topic compositions of \de and \zh \HS share some similarities. For instance, Military (e.g., battle reports) and Politics (e.g., authorities' policy and personnel changes) together account for more than half the stories in both languages. On the other hand, we also have language-specific observations. 9\% \de stories are about Literature (e.g., news about book publications), but this topic is not seen in \zh \HS. And while 14\% \de stories are about Sovereign (e.g., royal families and Holy See), there are only 2 examples in \zh (both about the emperor; we found no record on any religious leader in Wanli Gazette).  Also, the topics of Society (e.g., social events and judicial decisions) and Natural Disaster (e.g., earthquakes, droughts, and floods) are more prevalent in the \zh dataset.

\paragraph{Story length.}~~In news summarisation tasks, special attention is paid to the lengths of news stories and summaries (see Tab.~\ref{tab:dataset_stats}). Comparing \de \textsc{HistSumm} with the corresponding modern corpus \de MLSUM, we find that although historical news stories are on average 53\% shorter, the overall compression rate ($CR$s) is quite similar (6.8\% \textit{vs} 5.8\%), indicating that key points are summarised to similar extents. Following LCSTS~\citep{lcsts}, the table shows character-level data for \zh, but this is somewhat misleading. While most modern words are double-character, single-character words dominate the historical vocabulary, e.g., the historical word `\Chinese{朋}' (\textit{friend}) becomes `\Chinese{朋友}' in modern \zh. According to \citet{ratio=1.6}, this leads to a character length ratio of approximately 1:1.6 between parallel historical and modern samples. Taking this into account, the $CR$s for the \zh \textsc{HistSumm} and LCSTS datasets are also quite similar to each other.

When contrasting \de with \zh (regardless of historical or modern), we notice that the compression rate is quite different. This might reflect stylistic variations with respect to how verbose news reports are in different languages or by different writers.

\subsection{Vicissitudes of News}\label{ssec:evolution}
Compared with modern news, articles in \textsc{HistSumm} reveal several distinct characteristics with respect to writing style, posing new challenges for machine summarisation approaches.

\paragraph{Lexicon.} With social and cultural changes over the centuries, lexical pragmatics of both languages have evolved substantially~\citep{hist-research-2}. For \de, some routine concepts from hundreds of years ago are no longer in use today, e.g., the term `Brachmonat' (\textnumero 41), whose direct translation is \textit{fallow month}, actually refers to \textit{June} as the cultivation of fallow land traditionally begins in that month~\citep{german-cal}. We observe a similar phenomenon in \zh \HS, e.g., `\Chinese{贡市}' (\textnumero 24 and \textnumero 31) used to refer to markets that were open to foreign merchants, but is no longer in use. For \zh, additionally, we notice that although some historical words are still in use, their semantics have changed over time, e.g., meaning of `\Chinese{闻}'  has shifted from \textit{hear} to \textit{smell} (\textnumero 53), and that of `\Chinese{走}'  has changed from \textit{run} to \textit{walk} (\textnumero 25).

\paragraph{Syntax.} Another aspect of language change is that some historical syntax has been abandoned. Consider `\hlb{daß derselbe noch länger allda}/ \hly{biß der Frantz. Abgesandter von dannen widerum abreisen möge}/ \hlb{verbleiben soll}' (\textit{\hlb{the same should still remain there for longer}, \hly{until the France Ambassador might leave again}}) (\textnumero 33). We find the \hly{subordinate clause} is inserted within the \hlb{main clause}, whereas in modern \de it should be `\hlb{daß derselbe noch länger allda verbleiben soll}, \hly{biß der Frantz. Abgesandter von dannen widerum abreisen möge}'. For \zh, inversion is common in historical texts but becomes rare in the modern language. For example, sentence `\Chinese{王氏之\hlb{女}\hly{成仙}者}' (\textit{Ms. Wang's \hlb{daughter} \hly{who became a fairy}}) (\textnumero 65) where the \hly{attributive adjective} is positioned after the \hlb{head noun}, should be `\Chinese{王氏之\hly{成仙（的）}\hlb{女}}' according to modern \zh grammars. Also, we observe cases where historical \zh sentences without constituents such as subjects, predicates, objects, prepositions, etc. In these cases, contexts must be utilised to infer corresponding information, e.g., only by adding `\Chinese{居正}' (\textit{Juzheng}, a minister's name) to the context can we interpret the sentence `\Chinese{已，又为私书安之云}' (\textnumero 20) as \textit{`after that, (Juzheng) wrote a private letter to comfort him'}. This adds extra difficulty to the generation of summaries.

\paragraph{Writing style.}  To inform readers, a popular practice adopted by modern news writers is to introduce key points in the first one or two sentences~\citep{news-intro}. Many machine summarisation algorithms leverage this pattern to enhance summarisation quality by incorporating positional signals~\citep{edmundson1969, pg, gui-etal-2019-attention}. However, this rhetorical technique was not widely used in \textsc{HistSumm}, where crucial information may appear in the middle or even the end of stories. For instance, the keyword `Türck' (\textit{Turkish}) (\textnumero 33) first occurs in the second half of the story; in article \textnumero 7 of \zh \HS (see Tab.~\ref{tab:example}), only after reading the last sentence can we know the final outcome (i.e., the authority's seal had been saved from fire). 

\section{Methodology}\label{sec:method}

\begin{figure}[!t]
  \centering
  \includegraphics[trim={0 50 0 30}, clip, width=\columnwidth]{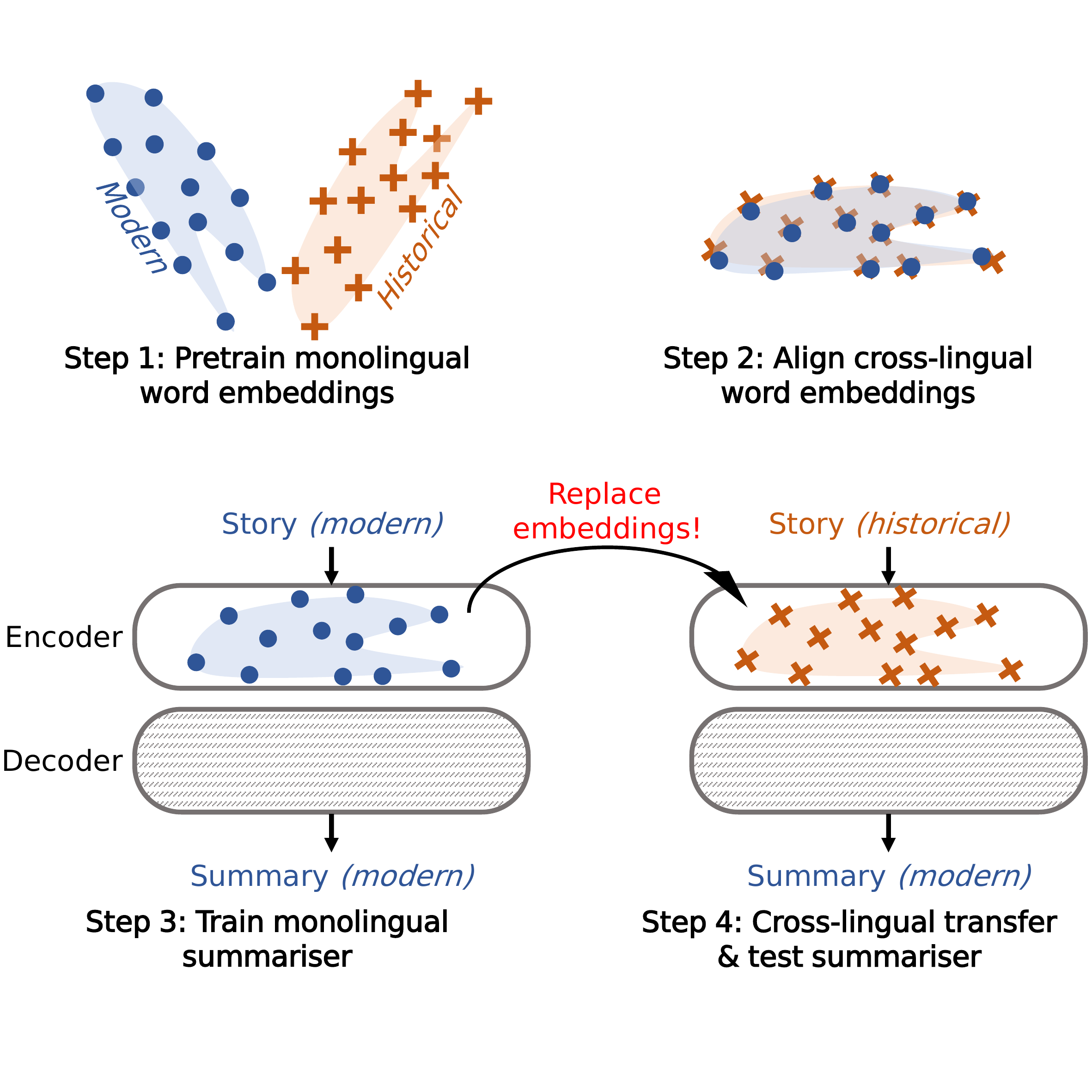}
\caption{Illustration of our proposed framework.}
\label{fig:pipeline}
\end{figure}

Based on the popular cross-lingual transfer learning framework of \citep{Ruder_survey}, we propose a simple historical text summarisation framework (see Fig.~\ref{fig:pipeline}), which can be trained even without supervision (i.e., parallel historical-modern signals).

\paragraph{Step 1.} For both \de and \zh, we begin with respectively training modern and historical monolingual word embeddings. Specially, for \de, following the suggestions of \citet{emb-eval}, we selected subword-based embedding algorithms (e.g., FastText~\citep{fasttext}) as they yield competitive results. In addition to training word embeddings on the raw text, for historical \de we also consider performing text normalisation (\textbf{\texttt{NORM}}) to enhance model performance. This orthographic technique aims to convert words from their historical spellings to modern ones, and has been widely adopted as a standard step by NLP applications for historical alphabetic languages~\citep{norm-app}. Although training a normalisation model in a fully unsupervised setup is not yet realistic, it can get bootstrapped with a single lexicon table to yield satisfactory performance~\citep{cSMTiser1, cSMTiser2}. 

For ideographic languages like \zh, word embeddings trained on stroke signals (which is analogous to subword information of alphabetic languages)  achieve state-of-the-art performance~\citep{cw2vec}, so we utilise them to obtain monolingual vectors. Compared with simplified characters (which dominate our training resources), traditional ones typically provide much richer stroke signals and thus benefit stroke-based embeddings~\citep{trad-better}, e.g., traditional `\TradChinese{葉}' (\textit{leaf}) contains semantically related components of `\Chinese{艹}' (\textit{plant}) and `\Chinese{木}' (\textit{wood}), while its simplified version (\Chinese{`叶'}) does not. 

Therefore, to improve the model performance we also conduct additional experiments on enhanced corpora which are converted to the traditional glyph using corresponding rules (\textbf{\texttt{CONV}}) (see \cref{ssec:model-conf} for further details). 

\paragraph{Step 2.} Next, we respectively build two semantic spaces for \de and \zh, each of which is shared by historical and modern word vectors. This approach, namely cross-lingual word embedding mapping, aligns different embedding spaces using linear projections~\citep{vecmap, Ruder_survey}. Given parallel supervision is very limited in real-world scenarios, we mainly consider two bootstrapping strategies: in a fully unsupervised (\textbf{\texttt{UspMap}}) style and through identical lexicon pairs (\textbf{\texttt{IdMap}}). While the former only relies on topological similarities between input vectors, the latter additionally takes advantage of words in the intersected vocabulary as seeds. Although their historical and current meanings can differ (cf. \cref{ssec:evolution}), in most cases they are similar, providing very weak parallel signals (e.g., `Krieg' (\textit{war}) and `Frieden' (\textit{peace}) are common to historical and modern \de; `\Chinese{天}' (\textit{universe}) and `\Chinese{人}' (\textit{human}) to historical and modern  \zh).

\paragraph{Step 3.} In this step, for each of \de and \zh we use a large monolingual modern-language summarisation dataset to train a basic summariser that only takes modern-language inputs. Embedding weights of the encoder are initialised with the \textit{modern} partition of corresponding cross-lingual word vectors in Step 2 and are frozen during the training process, while those of the decoder are randomly initialised and free to update through back-propagation.

\paragraph{Step 4.} Upon convergence in the last step, we directly replace the embedding weights of the encoder with the \textit{historical} vectors in the shared vector space, yielding a new model that can be fed with historical inputs but output modern sentences. This entire process does not require any external parallel supervision.

\section{Experimental Setup}

\subsection{Training Data}\label{ssec:datasets}
Consistent with \cref{ssec:annotation}, we selected \de MLSUM and \zh LCSTS as monolingual summarisation training sets. For monolingual corpora for word embedding training, to minimise temporal and domainal variation, we only considered datasets that were similar to articles in MLSUM, LCSTS, and \textsc{HistSumm}, i.e, with text from comparable periods and centred around news-related domains.

For modern \de, such resources are easy to access: we directly downloaded the \de News Crawl Corpus released by WMT 2014 workshops~\citep{wmt14}, which contains shuffled sentences from online news sites. We then conducted tokenisation and removed noise such as emojis and links.
For historical \de, besides the already included GerManC corpus, we also saved Deutsches Textarchiv~\citep{dta-data}, Mercurius-Baumbank~\citep{mb-data}, and Mannheimer Korpus~\citep{mann-data} as training data. 
Articles in these datasets are all relevant to news and have topics such as Society and Politics. Note that we only preserved documents written in 1600 to 1800 to match the publication time of \de \HS stories (cf. \cref{ssec:stats}). 
Apart from the standard data cleaning procedures (tokenisation and noise removal, as mentioned above), for historical \de corpora we replaced the very common slash symbols (/) with their modern equivalents: commas (,)~\citep{slash-comma}. We also lower-cased letters and deleted sentences with less than 10 words, yielding 505K sentences and 12M words in total.

For modern \zh, we further collected news articles in the corpora released by \citet{zh-news}, \citet{nlpcc}, and \citet{clue} to train better embeddings. For historical \zh, to the best of our knowledge, there is no standalone Ming Dynasty news collection except Wanli Gazette. Therefore, from the resources released by \citet{ming-data}, we retrieved Ming Dynasty articles belonging to categories\footnote{Following the topic taxonomy of \citet{ming-data}.} of Novel, History/Geography, and Military\footnote{Sampling inspection confirmed that their domains are similar to those of Wanli Gazette.}. 
Raw historical \zh text does not have punctuation marks, so we first segmented sentences using the Jiayan Toolkit\footnote{\url{https://github.com/jiaeyan/Jiayan}}. Although Jiayan supports tokenisation, we skipped this step as the accuracy is unsatisfactory. Given that a considerable amount of historical \zh words only have one character (cf. \cref{ssec:stats} and \cref{ssec:evolution}), following \citet{P18-2023} we simply treated characters as basic units during training. Analogous to historical \de, we removed sentences with less than 10 characters. The remaining corpus has 992k sentences and 28M characters.

\subsection{Baseline Approaches}

In addition to the proposed method, we also consider two strong baselines based on the Cross-lingual Language Modelling paradigm (XLM)~\citep{xlm}, which has established state-of-the-art performance in the standard cross-lingual summarisation task~\citep{jointly-acl20}. More concretely, for \de and \zh respectively, we pretrain baselines on all available historical and modern corpora using causal language modelling and masked language modelling tasks. Next, they are respectively fine-tuned on modern text summarisation and unsupervised machine translation tasks. The former becomes the (\textbf{\texttt{XLM-E2E}}) baseline, which can be directly executed on \textsc{HistSumm} in an end-to-end fashion; the latter (\textbf{\texttt{XLM-Pipe}}) is coupled with the basic summariser for modern inputs in Step 3 of \cref{sec:method} to form a translate-then-summarise pipeline.

\subsection{Model Configurations}\label{ssec:model-conf}
\paragraph{Normalisation and convention.} We normalised historical \de text using cSMTiser~\citep{cSMTiser1, cSMTiser2}, which is based on character-level statistical machine translation. Following the original papers, we pretrained the normaliser using RIDGES corpus~\citep{ridges}.
As for the \zh character convention, we utilised the popular OpenCC\footnote{\url{https://github.com/BYVoid/OpenCC}} project which uses a hard-coded lexicon table to convert simplified input characters into their traditional forms.

\paragraph{Word embedding.} As discussed in \cref{sec:method}, when training \de and \zh monolingual embeddings, we respectively ran subword-based FastText~\citep{fasttext} and stroke-based Cw2Vec~\citep{cw2vec}. For both languages, we set the dimension at 100 and learned embeddings for all available tokens (i.e., \texttt{minCount} = 1). Other hyperparameters followed the default configurations. After training, we preserved the most frequent 50K tokens in each vocabulary (NB: historical \zh only has 13K unique tokens). To obtain aligned spaces for modern and historical vectors, we then utilised the robust VecMap framework~\citep{vecmap} with its original settings.

\paragraph{Summarisation model.} We implemented our main model based on the robust Pointer-Generator Network~\citep{pg},
which is a hybrid framework for extractive (to copy source expressions via pointing) and abstractive (to produce novel words) summarisation models.
After setting up the encoder and decoder (cf. in Step 3 of \cref{sec:method}), we started training with the default configurations. As for the two baselines which are quite heavyweight (XLM~\citep{xlm} is based on BERT~\citep{bert} and has 250M valid parameters), we trained them from scratch with FP16 precision due to moderate computational power access. All other hyperparameter values followed the official XLM settings. To ensure the baselines can yield their highest possible performance, we trained them on the enhanced corpora, i.e., normalised \de (\texttt{NORM}) and converted \zh (\texttt{CONV}).

\section{Results and Analyses}

\subsection{Automatic Evaluation}\label{ssec:auto_eval}
\begin{table}[!t]\small
\centering 
\begin{tabular}{{p{2.4cm}P{1.7cm}P{1.7cm}P{1.7cm}}}
\toprule 
\de & ROUGE-1 & ROUGE-2 & ROUGE-L \\ \midrule
\texttt{XLM-Pipe} & 12.72 & 2.88 & 10.67 \\ 
\texttt{XLM-E2E} & 13.48 & 3.27 & 11.25 \\ \hdashline
\texttt{UspMap} & 13.36 & 3.02 & 11.28 \\ 
\texttt{UspMap}+\texttt{NORM} & 13.78 & \textbf{3.59} & 11.60 \\
\texttt{IdMap} & 13.45 & 3.10 & 11.38 \\  
\texttt{IdMap}+\texttt{NORM} & \textbf{14.37} & 3.30 & \textbf{12.14} \\ \midrule
\zh & \\ \midrule
\texttt{XLM-Pipe} & 10.91 & 2.96 & 9.83 \\ 
\texttt{XLM-E2E} & 12.67 & 3.86 & 11.02 \\ \hdashline
\texttt{UspMap} & 13.09 & 4.25 & 11.31 \\ 
\texttt{UspMap}+\texttt{CONV} & 16.38 & 6.06 & 14.00 \\
\texttt{IdMap} & 18.38 & 7.05 & 15.89 \\  
\texttt{IdMap}+\texttt{CONV} & \textbf{19.22} & \textbf{7.42} & \textbf{16.52} \\\bottomrule
\end{tabular}
\caption{ROUGE F1 scores (\%) on \textsc{HistSumm}.}
\label{tab:auto_eval}
\end{table}
\begin{table}[!t]\small
\centering 
\begin{tabular}{{p{2.4cm}P{1.7cm}P{1.7cm}P{1.7cm}}}
\toprule 
\textsc{en}$\rightarrow$\zh & ROUGE-1 & ROUGE-2 & ROUGE-L \\ \midrule
\texttt{XLM-Pipe} & 14.93 & 4.14 & 12.62 \\ 
\texttt{XLM-E2E} & \textbf{18.02} & \textbf{5.10} & \textbf{15.39} \\ 
\hdashline
\texttt{UspMap} & 11.43 & 1.27 & 10.07 \\ 
\texttt{IdMap} & 12.06 & 1.72 & 10.93 \\ \midrule
\textsc{zh}$\rightarrow$\en & \\ \midrule
\texttt{XLM-Pipe} & 9.08 & 3.29 & 7.43 \\ 
\texttt{XLM-E2E} & \textbf{12.97} & \textbf{4.31} & \textbf{10.95} \\ 
\hdashline
\texttt{UspMap} & 5.15 & 0.84 & 2.42 \\ 
\texttt{IdMap} & 5.98 & 1.33 & 2.90 \\\bottomrule
\end{tabular}
\caption{ROUGE F1 scores (\%) of \textit{standard} cross-lingual summarisation. Following \citet{jointly-acl20}, for monolingual pretraining, we used corpora in \cref{ssec:model-conf} (57M sentences) for modern \zh and annotated Gigaword~\citep{gigaword} (183M sentences) for \en; for summarisation training, we used LCSTS for \textsc{en}$\rightarrow$\zh and CNN/DM dataset~\citep{cnn/dm} for \textsc{zh}$\rightarrow$\en; for testing, we used the data released by \citet{multitask}.}
\label{tab:std_xlm}
\end{table}
\begin{table*}[!t]\small
\centering 
\begin{tabular}{{p{3.0cm}P{2.3cm}P{2.3cm}P{2.3cm}P{2.3cm}}}
\toprule 
\de & Informativeness & Conciseness & Fluency & Currentness  \\ \midrule
\texttt{\texttt{Expert}}  & 4.85 (.08) & 5.00 (.00) & 4.94 (.03) & 4.99 (.00) \\ \hdashline
\texttt{\texttt{XLM-E2E}}  & 2.26 (.20) & 2.35 (.24) & \textbf{3.34} (.19) & 3.67 (.23) \\ 
\texttt{UspMap}+\texttt{NORM}  & 2.51 (.18) & 2.53 (.22) & 3.28 (.22) & 3.64 (.24)  \\
\texttt{IdMap}+\texttt{NORM}  & \textbf{2.52} (.18) & \textbf{2.54} (.20) & 3.32 (.28) & \textbf{3.72} (.24) \\ \midrule
\zh & \\ \midrule
\texttt{\texttt{Expert}}  & 4.72 (.10) & 4.98 (.01) & 4.97 (.02) & 4.90 (.04) \\ \hdashline
\texttt{\texttt{XLM-E2E}}  & 2.18 (.23) & 2.21 (.27) & \textbf{2.80} (.22) & 2.53 (.23) \\ 
\texttt{IdMap}  & \textbf{2.39} (.19) & 2.49 (.26) & 2.66 (.25) & 2.50 (.23)  \\  
\texttt{IdMap}+\texttt{CONV}  & 2.37 (.21) & \textbf{2.57} (.28) & 2.78 (.24) & \textbf{2.59} (.25) \\\bottomrule
\end{tabular}
\caption{Average human ratings on \textsc{HistSumm} (variance is in parentheses).}
\label{tab:human_eval}
\end{table*}

We assessed all models with the standard ROUGE metric~\citep{rouge}, reporting F1 scores for ROUGE-1, ROUGE-2, and ROUGE-L. Following \citet{lcsts}, the ROUGE score of \zh outputs are calculated on character-level.

As shown in Tab.~\ref{tab:auto_eval}, for \de, our proposed methods are comparable to the baseline approaches or outperform the baselines by small amounts; for \zh, our models are superior by large margins. Given that XLM-based models require a lot more training resources than our model, we consider this a positive result. For comparison of the strengths and weaknesses of the models, we show their performance for a modern cross-lingual summarisation task in Tab.~\ref{tab:std_xlm}. To heighten the contrast we chose two languages (\zh and \en) from different families and with minimal overlap of vocabulary. As shown in Tab.~\ref{tab:std_xlm}, the XLM-based models outperform our method on this modern language cross-lingual summarisation task by large margins.

The difference in the performance of models on the modern and historical summarisation tasks illustrate key differences in the tasks and also some of the shortcomings of the models. Firstly, the great temporal gap (up to 400 years for \de and 600 years for \zh) between our historical and modern data hurts the XLM paradigm, which relies heavily on the similarity between corpora~\citep{umt-useless}. In addition, \citet{umt-useless} also show that inadequate monolingual data size (less than 1M sentences) is likely to lead to unsatisfactory performance of XLM, even for etymologically close language pairs such as \en-\de. In our experiments we only have 505K and 992K sentences for historical \de and \zh (cf. \cref{ssec:datasets}). On the other hand, considering the negative influence of the error-propagation issue (cf. \cref{sec:rw}), the poor performance of \texttt{XLM-Pipe} is not surprising and is in line with observations of \citet{jointly-acl20} and \citet{attend-acl20}.
Our model instead makes use of cross-lingual embeddings, including bootstrapping from identical lexicon pairs. This approach helps overcome data sparsity issues for the historical summarisation tasks and is also successful at leveraging the similarities in the language pairs. However, its performance drops when the two languages are as far apart as \en and \zh.

When analysing the ablation results of the proposed method, on \de and \zh we found different trends. For \de, scores achieved by all the four setups show minor variance. To be specific, models bootstrapped with identical word pairs outperformed the unsupervised ones, and models trained on normalised data yielded stronger performance. Among all tested versions, \texttt{UspMap}+\texttt{NORM} got the best score in ROUGE-2 and \texttt{IdMap}+\texttt{NORM} led in ROUGE-1 and ROUGE-L, indicating that the normalisation enhancement does benefit \de historical text summarisation models.
For \zh, as predicted, with richer glyph information encoded, the stroke-based embedding method can better learn word semantics. We find that \texttt{UspMap}+\texttt{CONV} outperforms \texttt{UspMap} and \texttt{IdMap}+\texttt{CONV} outperforms \texttt{IdMap}. Adding identical words during mapping initialisation brings substantial benefits too: 3.58\% and 2.52\% ROUGE-L improvement for \texttt{IdMap} over \texttt{UspMap} and \texttt{IdMap}+\texttt{CONV} over \texttt{UspMap}+\texttt{CONV}, respectively.

\subsection{Human Judgement}\label{ssec:exp-human}

To gain further insights, we invited six experts to conduct human evaluations. Like the annotators in \cref{ssec:annotation}, they also held degrees in Germanistik or Ancient Chinese Literature. Beyond the standard dimensions of summarisation evaluation (Informativeness, Conciseness, and Fluency), we added `Currentness' as the fourth, which focuses on measuring `to what extent a summary follows current rather than early linguistic styles'.
We used a five-point Likert scale, with 1 for worst and 5 for best. For each language, experts were only asked to rate the gold-standard human summary and the summaries generated by the \texttt{XLM-E2E} baseline and the best two setups in \cref{ssec:auto_eval}. For each of the 100 news stories in each language, 3 experts independently each rated the three model outputs and the human summary.

The final results are given in Tab.~\ref{tab:human_eval}. When comparing different systems, we report statistical significance as the $p$-value of two-tailed t-tests with Bonferroni correction~\citep{p-value}. We found that in all aspects the scores for the gold-standard summaries were always above 4 points, indicating the high quality of the gold-standard summaries.
Across both languages, our models outperform the baseline for informativeness and conciseness ($p$\textless0.01) and achieve comparable levels of fluency and currentness. Summaries generated by \texttt{XLM-E2E} were slightly more fluent than our approach for both \de and \zh ($p$\textless0.05), indicating that the baseline has merit with respect to its language modelling abilities. However, it tended to make errors in understanding historical inputs and locating key points; e.g. the human reference for \zh article \textnumero 57 is focused on the commander's decision of bursting the river to beat the rebel army (\Chinese{`宁夏之役中，魏学曾为了击溃叛乱部落，决定决河灌城'}), but \texttt{XLM-E2E} summarises it as \Chinese{黄河大堤水，比塔顶还高几丈}' (\textit{the surface of the river is several feet higher than the tower top}), which is fluent but irrelevant.

As for different setups of the proposed algorithm, for \de, in dimensions of Informativeness, Conciseness and Fluency, the performance of \texttt{UspMap}+\texttt{Norm} and \texttt{IdMap}+\texttt{NORM} was almost equally good. The improvement from utilising identical word pairs for cross-lingual word embedding mapping seems more evident for Currentness, i.e., the average score was 0.08 higher ($p$\textless0.05).
For \zh, while \texttt{IdMap} and \texttt{IdMap}+\texttt{CONV} achieved close Informativeness scores, the latter outperforms the former in other three aspects by 0.08, 0.12, and 0.09 respectively ($p$\textless0.01). This observation indicates that when the lexical encoding is improved with enriched stroke-level information, the model is less likely to include redundant information in the summaries (i.e., conciseness score is higher), and the produced sentences are more fluent in terms of modern \zh grammars (see output examples in Appendix~\ref{app:cases}).

\subsection{Error Analysis}

We further analysed model inputs with the lowest scores in \cref{ssec:exp-human}, and found that they were mostly for stories whose content was dissimilar to \textit{any} sample in modern training sets. For instance, five \zh texts in \HS are on themes not seen in modern news (i.e., witchcraft (\textnumero 65), monsters (\textnumero 35 and \textnumero 46), and abnormal astromancy (\textnumero 8 and \textnumero28)). On these texts, even the best-performing \texttt{IdMap}+\texttt{CONV} model outputs a large number of [UNK] tokens and can merely achieve average Informativeness, Conciseness, Fluency, and Correctness scores of 1.41, 1.67, 1.83, and 1.60 respectively, which are significantly below its overall results in Tab.~\ref{tab:human_eval}. 
This reveals the current system's shortcoming when processing inputs with theme-level \textit{zero-shot} patterns. This issue is typically ignored in the cross-lingual summarisation literature due to the rarity of such cases in modern language tasks.  However, we argue that a key contribution of our proposed task and dataset is that they together indicate new improvement directions beyond standard cross-lingual summarisation studies, such as the challenges of zero-shot generalisation and historical linguistic gaps (cf. \cref{ssec:evolution}).

\section{Conclusion and Future Work} 
This paper introduced the new task of summarising historical documents in modern languages, a previously unexplored but important application of cross-lingual summarisation that can support historians and digital humanities researchers. To facilitate future research on this topic, we constructed the first summarisation corpus for historical news in \de and \zh using linguistic experts. We also proposed an elegant transfer learning method that makes effective use of similarities between languages and therefore requires limited or even zero parallel supervision. Our automatic and human evaluations demonstrated the strengths of our method over state-of-the-art baselines. This paper is the first study of automated historical text summarisation. In the future, we will improve our models to address the issues highlighted in this study (e.g. zero-shot patterns and language change), add further languages (e.g., English and Greek), and increase the size of the dataset in each language.

\section*{Acknowledgements}
This work is supported by the award made by the UK Engineering and Physical Sciences Research Council (Grant number: EP/P011829/1) and Baidu, Inc.
Neptune.ai generously offered us a team license to facilitate experiment tracking.

We would like to express our sincerest gratitude to Qirui Zhang, Qingyi Sha, Xia Wu, Yu Hu, Silu Ding, Beiye Dai, Xingyan Zhu, and Juecheng Lin, who are all from Nanjing University, for manually annotating and validating the \HS corpus. 
We also thank Guanyi Chen, Ruizhe Li, Xiao Li, Shun Wang, Zhiang Chen, and the anonymous reviewers for their insightful and helpful comments.

\bibliography{eacl2021}
\bibliographystyle{acl_natbib}

\appendix
\onecolumn
\section{Output Samples}\label{app:cases}

\begin{small}
\centering 
\bgroup
\def\arraystretch{1.5}
\begin{longtable}{{p{2.7cm}p{12.5cm}}}
\toprule 
\de & \textnumero 11\\ \midrule
Story & Die Arbeiten im hiesigen Arsenal haben schon seit langer Zeit nachgelassen, und seitdem die Perser so sehr von den Russen geschlagen worden sind, hört man überhaupt nichts mehr von Kriegsrüstungen in den türkischen Provinzen. Die Pforte hatte nicht geglaubt, daß Rußland eine so starke Macht nach den Ufern des kaspischen Meeres abschicken, und daß der Krieg mit den Persern sobald eine so entscheidende Wendung nehmen würde. Alle kriegerischen Nachrichten, die wir jetzt aus den türkischen Provinzen erhalten, erstrecken sich blos auf die bewaffneten Räuber-Korps, die in der Gegend von Adrianopel noch immer ihren Unfug fortsetzen, der auch wohl nicht eher aufhören wird, bis die Pascha's selbst bestraft worden sind, die die Räuber beschützen. - Im Anfange dieses Monats erschien eine russische Fregatte am Eingange des schwarzen Meeres, ward durch Sturm vor den türkischen Forts vorbei in den Kanal getrieben, ohne daß die Kommandanten dieser Forts ihr den geringsten Widerstand entgegen stellen konnten, und legte sich, Bujukdere gegenüber, vor Anker. Sobald der Kapitän-Pascha dies erfuhr, verfügte er, daß jene Kommandanten abgesetzt werden sollten, und beschwerte sich bei dem hiesigen russischen Minister darüber, daß jenes Kriegsschiff sich unterstanden habe, wider alle Stipulationen der Traktaten in den Kanal einzulaufen. Nachdem aber der Zufall, wodurch dies geschehen ist, näher aufgeklärt war, widerrief der Kapitän-Pascha die Befehle, die gegen die Kommandanten der an dem Kanal gelegenen Forts erlassen wurden. Auch ward auf Ansuchen des russischen Gesandten der gedachten Fregatte aller mögliche Beistand geleistet, um sich repariren, und nach der Krimm, woher sie gekommen war, zurückkehren zu können. - Die Gesandten, welche die Pforte schon seit 2 Jahren nach Wien und Berlin bestimmt hat, sind noch immer hier; dies beweiset, daß alle Schwierigkeiten in Rücksicht dieser Missionen noch nicht gehoben sind Der nach Paris bestimmte türkische Gesandte wird aber, wie es heißt, bald abreisen. - Zwei sehr angesehene französische Offiziers, die in türkischen Dienst getreten waren, sind wieder aus demselben entlassen worden. \newline \textit{\textcolor{darkblue}{(The work in the arsenal has for a long time slacked off. And since the Persians were beaten so badly by the Russians, people have heard complete nothing about war armaments in the durkian provinces.
The Porte would not have thought that Russia would send such a powerful force to the shores of the Caspian Sea, and that the war with the Persians would at the same time take such a decisive turn.
All belligerent news, that we now receive from the Durkian provinces, extends only to the armed robber corps, which are in the area of Adrianopl still continuing their mischief, which is still unlikely to end until the pashas themselves, who protect the robbers, have been punished.
- At the beginning of this month a Russian frigate appeared at the entrance to the Black Sea, was driven by a storm past the Durkian forts into the channel, without that the commanders of this fort could oppose it with the slightest resistance, and (the Russian frigate) presented itself across from Bujukdere at anchor.
As soon as the captain Pasha found out about this, he decreed that those commanders should be deposed and complained to the local Russian minister about that that that grieg ship had dared to enter the canal, against all stipulations of the tracts.But after the coincidence, by which this happened, had been more clearly clarified, the captain-pasha recalled the orders, which would be enacted against the commanders of the fort on the canal.
Also, at the request of the Russian confession, the intended frigate was given all possible assistance in order to repair itself and to be able to return to Grimm, whence it had come.
- the confessions, that the gate has set for Vienna and Berlein for two years, are still here; this proves that all difficulties in regard to these missions have not yet been resolved, but the destined-for-Paris Durkian legate will, as it is said, soon be leaving.
- two very highly respected French officers, who had entered Durk service, have been dismissed from the same.)
}} \\ \hdashline
\texttt{Expert}  & Wie es zwischen Russland und der Türkei lief, war noch unsicher. \newline \textit{\textcolor{darkblue}{(How things would go between Russia and Turkey, was still uncertain.)}} \\ 
\texttt{IdMap}+\texttt{NORM} & die [unk] des [unk] zeigen , dass der krieg mit den persern sobald eine so entschiedende wendung nehmen würde . die wendung eines blauen wunders ist nicht nur zu sehen , wie man es weitergeht . \newline \textit{\textcolor{darkblue}{([unk] show that the war with the Persians would very soon take such a decisive turn. The turning point of a blue miracle is not just to see how it goes on.) }} \\ 
\texttt{UspMap}+\texttt{NORM} & die arbeiten im arsenal haben schon seite länger zeit nachgelassen , und seitedem die perser so sehr von den russen geschlagen worden sind , hört man überhaupt nichts mehr von kriegrüstungen in den durkischen provinzen . \newline \textit{\textcolor{darkblue}{(The work in the arsenal has for a long time slacked off. And since the persians were beaten so badly by the russians, people have heard complete nothing about war armaments in the durkian provinces.)}} \\ 
\bottomrule
 \pagebreak

\toprule
\de & \textnumero 33\\ \midrule
Story & ES befindet sich schon etliche Tage hero von der Crone Schweden ein Abgeordneter incognito allhier/ aber noch unbewust in was Negotio. Am verwichenen Montage ist von Ihrer Käyserl. M. an den Abgesandten zu München Herrn Grafen von Königsegg ein Currirer abgeschickt worden/ wie man vernimt/ weilen I. Chur-Fürstl. Durchl. allda gegen I. Käyserl. M. hoch contestirt/ daß Sie Dero Käyserl. und Röm. Reichs Intereße auff alle möglichste Weise befördern helffen/ und auff solche Resolution gedachter Kayserl. Abgesandter von dar seine Reise in auffgetragener LegationsCommißion weiters nehmen wollen/ daß derselbe noch länger allda/ biß der Frantz. Abgesandter von dannen widerum abreisen möge/ verbleiben soll/ damit I. Chur-Fürstl. Durchl. durch erstgedachten Abgesandten nicht zu andern Gedancken kommen möchte. Vorgestern ist der Käys. neulich zu dem Vezier nacher Ofen geschickte Türck. Ober-Dolmetscher/ Herr Minnisky wider zurücke anhero gekommen/ von welchen man vernimt/ daß gedachter Vezier/ wie auch die Baßen von Erlau und Waradein/ sich wegen des beschuldigten Unterschleiffs der Rebellen sehr excusirt/ und negirt/ daß sie bißhero ihrem gethanen Versprächen zu wider die Rebellen in ihren Territoriis wißentlich geduldet hätten/ sondern solches vil mehrers von dem Abassy geschehen wäre/ und habe gedachter Vezier sein hievoriges Versprächen gegen I.K.M. nochmal höchstens contestiren laßen: Demnach aber/ ungeachtet diser Sinceration/ man gewiß weiß/ daß obgedachte Rebellen nicht allein von den Türcken in ihren Gebieten geduldet/ sondern auch bewaffnet worden/ und in neulicher Action die Türcken auff Seiten der Rebellen selbsten darbey gewesen/ also läst es sich nun zu einer würcklichen Ruptur ansehen/ deßwegen auch bey Hofe vil Patenten auff neue Werbungen heraus gegeben werden. \newline \textit{\textcolor{darkblue}{(A few days ago there was a member of parliament incognito here from the Royal Family of Sweden, but still unconsciously in some business. On the elapsed Monday, a Currier was sent from their Royal M to the emissaries to monks, Grafen von Königsegg, as people hear, that I. Chur-Fürstl Durchl is contesting against I. Royal M, that they help to promote the Royal and Roman Empire interests in every possible way, and that Royal Abgesander who is thinking of such a resolution, wants to continue his journey in the applying Legations Commission, that the same should still remain there for longer, until the Franz Abgesander might leave again, so that I. Chur-Fürstl Durchl through the first envisaged delegate does not want to come to other thoughts. The day before yesterday Käy's new Türck interpreter, Mr. Miniski, who was sent to the Vezier afterwards, has come here, from whom people heard that the intended Vezier, like the bases of Erlau and Waradien, were for the accused hiding of the rebels very excited, and denied that they had so far knowingly tolerated their promise against the rebels in their territories, but that such a thing would have happened much more from the Abassi, and thought Vezier had made his previous promise against the IKM. at most let them contest again: but regardless of this sinceration, people know for sure, the contemplated rebels are not only tolerated by the Tirken in their areas, but also been armed, and in the recent action the Turks were themselves there on the side of the rebels, so it can be viewed now as a real rupture, which is why at court many patents on new recruitments are issued.) }} \\ \hdashline
\texttt{Expert} & Der Kaiser versuchte, durch Verhandlungen seine Interessen gewährzuleisten. Inzwischen boten die Türken wider Versprechen den Rebellen Unterstützung \newline \textit{\textcolor{darkblue}{(The emperor tried to safeguard his interests through negotiations. Meanwhile Turks broken the promise and provided support to the rebels.)}} \\ 
\texttt{IdMap}+\texttt{NORM} & es befindet sich schon etliche tage her von der crone schweden ein abgeordneter inconitum allhier , aber noch unbewusst in was negotio . am verwichenen montage ist von ihrer käyserl . allda gegen i . \newline \textit{\textcolor{darkblue}{(A few days ago there was a member of parliament incognito here from the Royal Family of Sweden, but still unconsciously in some business. On the elapsed Monday is from their Royalty all against I.) }} \\ 
\texttt{UspMap}+\texttt{NORM} & es befindet sich schon etliche tage her von der crone schweden ein abgeordneter inconitum allhier , aber noch unbewusst in was negotio . am verwichenen montage ist von ihrer käyserl . m . an den abgestanden \newline \textit{\textcolor{darkblue}{(A few days ago there was a member of parliament incognito here from the Royal Family of Sweden, but still unconsciously in some business. On the elapsed Monday is from their Royal M to the stale ...)}} \\ \bottomrule

\pagebreak

\toprule
\de & \textnumero 34\\ \midrule
Story & Jhre Königl. Majest. befinden sich noch vnweit Thorn/ vnd seynd Cosakische Deputirte vnter Wegs/ jhr factum bey Seiner Majest. zu justificiren, vnd wegen jhrer Treu Versicherung zu thun. Von den Fridens. Tractaten zwischen Pohlen vnd Schweden ist noch wenig zu melden. Seithero die Pohlen bey Marienburg den Schweden eine Schantz/ der Kessel genant/ Abgenommen/ ist nichts weiters vorgefallen/ auch hiesiger Stadt Völcker vor dem Haupt noch nichts tentirt, jedoch sagt man daß noch dise woche etwas vorgehen werde/ so bald nur alle Battereyen in den 3 Quartieren fertig/ vnd die Mörser darauff gebracht worden/ vmb solches mit Feur zu bezwingen/ weil mit dem Schiessen doch nichts zugewinnen/ vnd der Sturm vnmöglich zu wagen ist/ daß aber das Brau- vnd Proviant Hauß darin in brand geschossen/ vnd die darinnen befindliche Cavallerie also ruinirt werden/ daß sie keinen Außfall mehr thun können/ ist gewiß/ deßgleichen hat der Obriste Zaphlizky mit 2000. Mann den Elbingern daß Viehe weggetriben/ welche darauff mit 500. Mann außgefallen/ solches wider zu erobern/ seynd aber mehrentheils nidergemacht/ vnd 6. vornehme Officierer neben vielen Gemeinen gefangen worden. So ist auch auß Churland über Memmel sichere Zeitung einkommen/ daß Herr General Duglas nur 2000. Mann nach Liffland gebracht/ vnd Pautzke sich mit Accord an die Pohlen ergeben habe/ seynd also von den Schweden in Mittau noch 300. Mann übrig/ deren Ergebung man nechstens zu verenehmen hoffet/ zumahlen selbige formaliter belägert seynd/ vnnd keinen Succurs zuvermuthen. Den gefangnen Hertzogen von Churland haben die Schweden wider in Liffland nach Revel gebracht/ dahero zu seiner Erledigung Hoffnung gemacht werden will. \newline  \textit{\textcolor{darkblue}{(Their Royal Majesties are still not far from Torn, and there are Cossack deputies on the way to their factum to be judged by His Majesty, and to be insured for their loyalty. (and for their loyal insurance to do that.
from the Fridens.
Tracts between Poland and Sweden are still little to be reported. Since the Pohlen near Marienburg took away a Schanz, which is called “boiler”, from the Swedes, nothing further has happened, also local city peoples have yet in the first place tented nothing, however, they say that something will happen this week,
as soon as all batteries are in the 3 quarters ready, and the mortars were brought to it, in order to defeat it with fire, because by shooting nothing could be gained, and the storm is impossible to be venture, but the brown-known and provisions house was set on fire and in it the cavalry were so ruined that they could no longer do any sorties, is certain.
Likewise, Colonel Zaplizki has driven away the cattle from the Elbingers with 2,000 men, who with 500 men failed to conquer such, but were mostly killed, and 6 distinguished officers were captured alongside many common ones.
It is also sure to be a newspaper from Churland via Memmel coming in, that General Duglas only brought 2000 men to Lifland, and Pauzke has surrendered to the Poles by accord, so from the Swedes in Mittau are still 300 men left, whose surrender is the next that people hoped to hear, as they are formally besieged, and no succurs can be expected. The Swedes have brought the captured Duke of Churland back to Revel in Lifland, therefore desired for his completion hope to be made.)
}} \\ \hdashline
\texttt{Expert}  & Der Krieg zwischen Polen und Schweden dauert an. Von einem Friedensvertrag ist noch nicht der Rede. \newline \textit{\textcolor{darkblue}{(The war between Poland and Sweden continues. Of the peace treaty is there still no talk.)}} \\ 
\texttt{IdMap}+\texttt{NORM} & ihre königl . maiest . befinden sich noch unweit toren , und sind cosakische deputierte unter weg , ihr factum bei seiner maiest . zu justifizieren , und wegen ihrer treu versicherung zu tun . \newline \textit{\textcolor{darkblue}{(Their Royal Majesties are still not far from Torn, and there are Cossack deputies on the way to their factum to be judged by His Majesty, and for their loyal insurance to do that.)}} \\ 
\texttt{UspMap}+\texttt{NORM} &  ihre königl . maiest . befinden sich noch unweit toren , und sind cosakische deputierte unter weg , ihr factum bei seiner maiest . zu justifizieren , und wegen ihrer treu versicherung zu tun . von den fridens . \newline \textit{\textcolor{darkblue}{(Their Royal Majesties are still not far from Torn, and there are Cossack deputies on the way to their factum to be judged by His Majesty, and for their loyal insurance to do that.)}} \\ 
\bottomrule

\pagebreak

\toprule
\de & \textnumero 39\\ \midrule
Story & Heute ist der Kayserl. General-Kriegs-Commissarius, Graf von Nesselrode, mit dem wegen Anweisung derer künftigen Winter-Quartiere abgefasseten Plan von Wien nach dem Kayserl. Haupt Quartier Haydelberg, zu des Printzen Eugenii Hoch-Fürstl. Durchl. wieder zurücke gegangen. Es verlautet dabey, daß die würckliche Einrichtung dererselben viele Schwürigkeiten gefunden habe, und daß verschiedene Reichs-Stände dieselbe von ihren Landen zuförderst abwenden wollen. Mehrere und besondere Umstände sind davon noch nicht bekand. An dem Kayserl. Hofe ist zu Bestreitung derer fortdaurenden schweren Kriegs-Kosten, beschlossen worden, auf verschiedene Waaren, und insonderheit auf den Wein u. Fische einen neuen Impost zu legen, ob aber auch künfftig das Silber-Geschirre in die Kayserl. Müntze dürfte gefordert werden, wie bisher verlauten will, solches ist noch zweyfelhafftig, inzwischen wird mit Eintreibung eines sogenandten Subsidii präsentanti, wobey alle vermögende Leute zur Anticipation eines nach eines jedweden Vermögen eingerichteten Quanti angehalten werden, und dargegen aus der Kayserl. Banco in 3. Jahren zahlbare Banco-Obligationen, nebst 5. pro Cent Interesse erhalten, nicht nur zu Wien fortgefahren, sondern es soll auch dergleichen in allen Kayserl. Erb-Landen, das eintzige Königreich Ungarn ausgenommen, dessen Privilegia solches nicht verstatten, eingeführet werden. Man hat aus Italien Nachricht, daß die Alliirten ten zwischen der Etsch und Adige nicht nur eine starcke Linie gezogen haben, um denen Deutschen den Rück-Weg nach dem Mantuanischen gäntzlich zu benehmen, sondern sich auch gegen das Triedentinische ziehen, und daselbst einbrechen wollen. Sonst weiß man, daß der Erb-Printz und numehro regierende Durchl. Hertzog, Carl, von Braunschweig-Wolfenbüttel, die gesuchte Veniam ätatis von Ihro Kayserl. Maj. auf das erstattete Reichs-Hofe-Raths-Gutachten erhalten habe. \newline \textit{\textcolor{darkblue}{(Today the Lord General-Krügs-Commissarius, Count of Nesselbrote, is with with the plan, which is drawn up according to the instructions for their future winter quarters, from Vienna to the emperor's headquarter, Heidelberg, to the Prince Eugenii Hoch-Fürstl Durchl. Again, it passed back. It is said that the real institution of the same has found many difficulties, and that the various imperial estates want to turn them away from their lands first of all. Several and special circumstances are not yet known about. At the Royal Court it was decided to pay for the continuing heavy Krügs costs, on various goods, and in particular on the wine and the Fish a new tax to put, but whether in the future the silver dishes in the Royal Coin should be required, as has been so far announced, this is still doubtful. In the meantime with the collection of a so-called Subsidium Presentanti, (in which all wealthy people are encouraged to anticipate a Quanti set up according to one’s every wealth, and on the other hand from the Royal Bank received in 3 years payable bank debts, plus 5th interest per cent,) not only carried on in Vienna, but the same should be introduced in all Royal Hereditary Lands, the only kingdom of Hungary, with the exception of whose privileges do not permit such. People had news from Italy that the Allies had not only drawn a strong line between the Etsch and the Adige in order to deprive the Germans of the way back to Mantuan entirely, but also oppose the Tridentine and want to break in there. Otherwise knowing people that the Hereditary Prince and now ruling Highness Duke, Carl, von Braunschweig-Wolfenbutel, who have sought for Veniam ätatis from their Royal May,  have received the submitted Reichs-Hofe-RAts report.)}} \\ \hdashline
\texttt{Expert}  & Der Kaiser ließ einrichtungsbezogene, finanzielle, militärstrategische und personelle Anordnungen vornehmen, um den Krieg weiterzuführen. \newline \textit{\textcolor{darkblue}{(The emperor had ordered to make facility-related, financial, military-strategic and personnel arrangements in order to continue the war.)}} \\
\texttt{IdMap}+\texttt{NORM} & heute ist der kaiserl . general-krügs-commissarius , graf von nesselbrote , mit dem wegen angeweisung derer künftigen winter-quartiere abgefassten plan von wie nach dem kaiserl . haupt quartier heidelberg , zu des prinzen eugenii hoch-fürstl . \newline \textit{\textcolor{darkblue}{(Today the Lord General-Krügs-Commissarius, Count of Nesselbrote, is with with the plan, which is drawn up according to the instructions for their future winter quarters, from Vienna to the emperor's headquarter, Heidelberg, to the Prince Eugenii Hoch-Fürstl.)
 }} \\ 
\texttt{UspMap}+\texttt{NORM} & heute ist der kaiserl . general-krügs-commissarius , graf von nesselbrote , mit dem wegen angeweisung derer künftigen winter-quartiere abgefassten plan von wie nach dem kaiserl . haupt quartier heidelberg , zu des prinzen eugenii hoch-fürstl . \newline \textit{\textcolor{darkblue}{(Today the Lord General-Krügs-Commissarius, Count of Nesselbrote, is with with the plan, which is drawn up according to the instructions for their future winter quarters, from Vienna to the emperor's headquarter, Heidelberg, to the Prince Eugenii Hoch-Fürstl.)}} \\ 
\bottomrule

\pagebreak

\toprule
\de & \textnumero 50\\ \midrule
Story & Donau-Strohm vom 13. Weinm. Aus Breßlau hat man unterm 3. dieses folgende Nachricht: Vorgestern sind ungemein grosse Heere Heuschrecken über hiesige Stadt gezogen, deren Flug von 10. Uhr des Mittags bis gegen 4. Uhr Abends gedauret. Eine Colonne nahme bey nahem die gantze Breite der Stadt ein, und die Höhe betrug ohngefehr 130. bis 140. Ellen. Noch viele andere Colonnen breiteten sich in grosser Menge aus, und man berichtet aus Zotten, daß sie allda ebenfalls in grosser Menge durchgeflogen seyen. Dieses Ungeziefer verliehret auf seinem Marsch viele von seinen Cameraden, welche von den Krähen, Raben, Dohlen und andern Vögeln fleißig gefangen werden, welche den Bauch eines Heuschrecken samt dem Eingeweyde fressen, und das übrige auf die Erde fallen lassen, von denen man viele auf hiesigen Feldern gesehen. Gestern sind wiederum neue Schwärme hier ankommen, welche sich aber nicht gelagert, sondern, wie die andern, ihren Flug weiter genommen haben, und dieser ihr Zug dauret so lang, als lang die Sonne hell und warm scheinet. Auf die Nacht erhobe sich ein hefftiger Wind, der unsere bisherige warme Lufft ziemlich abgekühlet, weswegen wir heute wenig Heuschrecken sehen. Auf denen Gütern des Grafen von Schweidnitz, zu Stephansdorff, ohngefehr 4. Meilen von hier, hat dieses Ungeziefer grossen Schaden gethan, da dasselbe alle Wayde für das Vieh abgefressen, und vorgestern ist ein anderes unbeschreiblich starckes Heer über gedachte Güter gezogen, welches seinen Flug gegen Prochwitz und Liegnitz genommen. \newline \textit{\textcolor{darkblue}{(Danube stream from the 13th Weinm. From Wroclaw comes under the 3rd day the following message: The day before yesterday, a great number of locusts have flown over the local city, and their flight lasted from 10 a.m. to around 4 p.m. A column took up almost the whole breadth of the city, and the height was about 130 to 140 cubits. Also, many other columns spread out in great numbers, and it is reported according to the villi that they had also flown through there in great numbers. This vermin lost on its march many of its companions, who were by the crows, ravens, jackdaws, and other birds busily caught, which eat the belly of a locust and its entrails, and let the rest of them fall to the ground, many of which have been seen on local fields. Yesterday again new swarms have arrived here, who didn't know what to do, but, like the others, continued their flight, and this migration lasts as long as the sun shines bright and warm. In the night rose up a violent wind, which cooled down our previous warm air quite a bit, which is why we today see few locusts. On the property of the Count of Schweidnitz, at Stephandarf, about four miles from here, this vermin has done great damage, since it has eaten up all the woad for the cattle, and the day before yesterday has another indescribably strong army marched over intended goods, which its flight took against Prochwitze and Lignitz.) }} \\ \hdashline
\texttt{Expert} & eine große Menge von Heuschrecken sind durchgeflogen. Ihre Anzahl ist wegen der insektfressenden Vögel und hefttigen Wind gesunken. \newline \textit{\textcolor{darkblue}{(A great number of locusts have flown through. Their numbers have decreased due to insectivorous birds and violent winds.)}} \\ 
\texttt{IdMap}+\texttt{NORM} & donaunknownstrom vom 13 . weinem . aus breslau hat man unterm 3 . dieses folgende nachricht : vorgestern sind ungemein grosse her heuschrecken über hiesige stadt gezogen , deren flug von 10 . ihr des mittages bis gegen 4 . ihr abends gedauert . \newline \textit{\textcolor{darkblue}{(Danube stream from the 13th Weinm. From Wroclaw comes under the 3rd day the following message: The day before yesterday, a great number of locusts have flown over the local city, and their flight lasted from 10 a.m. to around 4 p.m. Her evening lasted.)}} 
\\ 
\texttt{UspMap}+\texttt{NORM} & donaunknownstrom vom 13 . weinem . aus breslau hat man unterm 3 . dieses folgende nachricht : vorgestern sind ungemein grosse her heuschrecken über hiesige stadt gezogen , deren flug von 10 . ihr des mittages bis gegen 4 . \newline \textit{\textcolor{darkblue}{(Danube stream from the 13th Weinm. From Wroclaw comes under the 3rd day the following message: The day before yesterday, a great number of locusts have flown over the local city, and their flight lasted from 10 a.m. to around 4 p.m.)}} \\ 
\bottomrule

\pagebreak

\toprule
\zh & \textnumero 27\\ \midrule
Story &  \Chinese{陕西巡抚叶詹熊奏称，三月初六日，黄色蔽天白昼黑暗,徐而变为红色，天鼓大鸣。初七日夜，从四广空降火一块，如盆大，后主三尾，落西北方。又查自十九年至今，天鼓时鸣，地道数震，火光冰雹，暴雨飓风，青气吐于兽吻，白气横于斗傍，彗自天飞，血从地涌。海内人民惊讶，以为从来无此非常之变，而且多也，疏上报闻。} \newline \textit{\textcolor{darkblue}{(Shaanxi Governor Ye Zhanxiong reported, on March 6th, a yellow colour covered the sky and the day was dark, it slowly turned red, thundered loudly. On the night of the seventh day in the lunar new year, a fire broke down from the sky, big as a basin, with three tails at the back, landed in the northwest. It was discovered that from the 19th year to now, it sometimes thunder, there were many earthquakes, fire and hail, rainstorms and hurricanes, black gas came out of animals' lips, white gas crisscrossed near buckets, comets flew in the sky, blood gushed from the ground. People all over the country were surprised, there had never been such unusual changes, and in such great number, the report was sent to the emperor.)}} \\ \hdashline
\texttt{Expert}  & \Chinese{陕西天鼓轰鸣，此现象长久以来时有发生，伴随有各类地质、气象和天文灾害。} \newline \textit{\textcolor{darkblue}{(It thundered in Shaanxi, this phenomenon has occurred from time to time for a long time, accompanied by a variety of geology, meteorological and astronomical disasters.) }} \\ 
\texttt{IdMap} & \Chinese{海内人民惊讶天鼓大鸣[UNK]。} \newline \textit{\textcolor{darkblue}{(People all over the country were surprised at thunder [UNK].)}} \\ \
\texttt{IdMap}+\texttt{CONV} &  \Chinese{陕西天鼓鸣[UNK]。} \newline \textit{\textcolor{darkblue}{(It thundered in Shaanxi [unk].)}} \\ \bottomrule

\\

\toprule
\zh & \textnumero 34\\ \midrule
Story &  \Chinese{宋应昌撤兵自朝鲜回京。兵部复预防计御应敌之机，以便夺责，以保万安等事。有旨，犬兵画撤宋应昌既难展布。着兴李如松取妄诈，未可遂称事完，督抚遥制不便，顾豢谋等谋素闲，着星夜前往彼处料理，务保成功，写敕与他，蓟镇防掳事务，暂令顺天边抚代管，宋应昌仍候顾豢。谦至日，交代回京。} \newline  \textit{\textcolor{darkblue}{(Song Yingchang withdrew his troops from Korea and returned to the capital. The Ministry of war has make new arranges to defend the enemies, in order to investigate the responsibilities, just in case. The emperor announces that, since Song Yingchang has trouble with his military deployment, Li Rusong is now commanded to fight with the arrogant enemies; the governor's remote control is inconvenient, and he is taking care of animals whilst doing other simple things; travel to the place overnight to handle it, be sure to succeed; write him orders, as for the war affairs in Jizhen, Shuntian governor is temporary in charge, Song Yingchang is relieved. On the next day, perform the handover and return to the capital.) }} \\ \hdashline
\texttt{Expert}  & \Chinese{宋应昌自朝鲜撤兵，兵部奉旨命李如松前去坐镇，交接成功后宋应昌方可回京。} \newline \textit{\textcolor{darkblue}{(Song Yingchang withdrew his troops from Korea, the Ministry of War ordered Li Rushong to go and take charge, after the handover Song Yingchang could return to the capital.) }} \\ 
\texttt{IdMap} & \Chinese{宋应昌撤兵自朝鲜回京。} \newline \textit{\textcolor{darkblue}{(Song Yingchang withdrew his troops from Korea and returned to the capital.)}} \\
\texttt{IdMap}+\texttt{CONV} &  \Chinese{宋应昌撤兵自朝鲜回京。}\newline \textit{\textcolor{darkblue}{(Song Yingchang withdrew his troops from Korea and returned to the capital.)}} \\ \bottomrule

\\

\toprule
\zh & \textnumero 70\\ \midrule
Story &  \Chinese{高拱卒。拱妻张氏上疏，陈乞恤典。上日：“高拱不忠，欺侮朕躬，今已死了，他妻还来乞恩，不准他。”居正言:“拱侍先帝潜邸，讲读有旧恩，宜予。”命复原职，给予祭葬。} \newline \textit{\textcolor{darkblue}{ (Gao Gong died. His wife Zhang wrote to the emperor, asking for a court ceremony. The emperor said: "Gao Gong was unfaithful, and intimidated me. Now he is dead, his wife comes back to ask for a favour, refuse." Zhang Juzheng said: "Gong served in the emperor's residence, there was past gratitude, ought to grant." It was ordered to resume Gong's original job, and grant him a funeral.) }} \\ \hdashline
\texttt{Expert}  & \Chinese{高拱去世，其妻上书求恩典遭到拒绝，经居正劝说后得到恩赐。} \newline \textit{\textcolor{darkblue}{(Gao Gong died, his wife Zhang wrote to the emperor asking for a court ceremony but she was rejected, after Juzheng's persuasion Gong was bestowed.) }} \\
\texttt{IdMap} & \Chinese{高拱不忠，已死了，他妻还来乞恩，不准他。} \newline \textit{\textcolor{darkblue}{(Gao Gong was unfaithful, he is dead, however his wife comes back to ask for a favour, refuse.)}} \\ 
\texttt{IdMap}+\texttt{CONV} &  \Chinese{高拱不忠，不准他妻来乞恩。} \newline \textit{\textcolor{darkblue}{(Gao Gong was unfaithful, don't allow his wife to come and ask for favour.) }} \\ \bottomrule

\pagebreak

\toprule
\zh & \textnumero 78\\ \midrule
Story &  \Chinese{黄台吉裴封顺义王。礼部等部尚书等官徐学谟等题称，北虏求嗣封爵，称黄台吉傈俺答嫡长男，应嗣王号，舍力克台吉傈黄台吉的男，应袭龙湖将军职衔。报可。} \newline \textit{\textcolor{darkblue}{(Huangtaiji was granted the title of King Shunyi. Xu Xuemo and others who were officials of the Ministry of Rites said, enemy from the north asked to be offered hereditary peerages, Huangtaiji was Anda's eldest son, he should inherit the title of king, Sheliketaiji was Huangtaiji's eldest son, he should inherit the rank of general Longhu. The suggestion was approved.) }} \\ \hdashline
\texttt{Expert}  & \Chinese{俺答部嫡长子黄台吉被礼部封顺义王。} \newline \textit{\textcolor{darkblue}{(Anda's eldest son Huangtaiji was granted the title of King Shunyi by the Ministry of Rites.) }} \\ 
\texttt{IdMap} & \Chinese{求嗣封爵，黄台吉傈俺答嫡长男，你知道吗？} \newline \textit{\textcolor{darkblue}{(Asking to be offered hereditary peerages, Huangtaiji was Anda's eldest son, do you know?)}} \\ 
\texttt{IdMap}+\texttt{CONV} &  \Chinese{黄台吉傈俺答嫡长男应袭龙湖将军职衔。} \newline \textit{\textcolor{darkblue}{(Huangtaiji was Anda's eldest son and should inherit the rank of general Longhu.) }} \\ \bottomrule

\\

\toprule
\zh & \textnumero 96\\ \midrule
Story &  \Chinese{山西宁武关军师作乱。军士李现等纠众三百，拥入兵备邢玠道门，逼挟粮米，鼓噪作乱。} \newline  \textit{\textcolor{darkblue}{(The troop in Ningwuguan Shanxi rioted. Sergeant Li Xian and other three hundred soldiers gathered, they entered the gate of Xingjiedao, robbed grain and rice, and clamoured to riot.) }} \\ \hdashline
\texttt{Expert} & \Chinese{宁武关军队以李现为首出现骚乱。} \newline \textit{\textcolor{darkblue}{(The troop in Ningwuguan had an riot which was lead by Lixian.) }} \\ 
\texttt{IdMap} & \Chinese{军师作乱：逼挟粮米，逼挟粮米，逼挟粮米，逼挟粮米，逼挟粮米，鼓噪作乱！} \newline \textit{\textcolor{darkblue}{(Troop rioted: robbed grain and rice, robbed grain and rice, robbed grain and rice, robbed grain and rice, robbed grain and rice, clamoured to riot!)}} \\
\texttt{IdMap}+\texttt{CONV} &  \Chinese{山西宁武关军师作乱。} \newline \textit{\textcolor{darkblue}{(The troop in Ningwuguan Shanxi rioted.)}} \\ \bottomrule
\end{longtable}
\egroup
\end{small}

\pagebreak
\section{Sample of Wanli Gazette (Scanned) Copies}\label{app:wanli}
\begin{figure}[!ht]
  \centering
  \includegraphics[width=\textwidth]{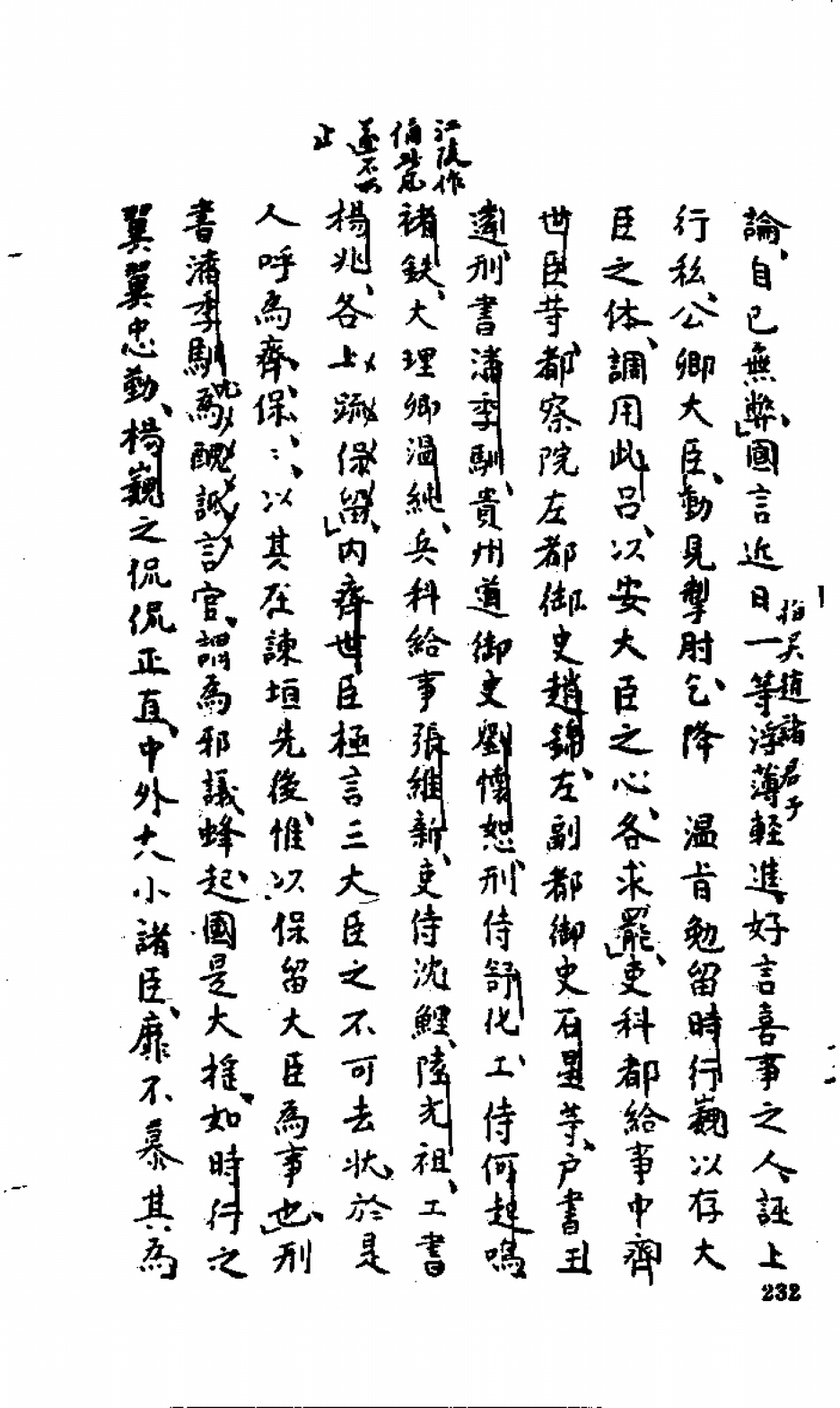}
\end{figure}

\end{document}